\documentclass[10pt,twocolumn,letterpaper]{article}

\usepackage{cvpr}              
\definecolor{cvprblue}{rgb}{0.21,0.49,0.74}
\usepackage[pagebackref,breaklinks,colorlinks,allcolors=cvprblue]{hyperref}

\usepackage{tabularx}
\usepackage{multirow}
\usepackage{balance}

\def\paperID{*****} 
\def\confName{CVPR}
\def\confYear{2026}

\title{SVBench: Evaluation of Video Generation Models on Social Reasoning}

\author{
  Wenshuo Peng$^{1}$~\footnotemark[1] ~\footnotemark[3]\quad
  Gongxuan Wang$^{3,4}$~\footnotemark[1]\quad
  Tianmeng Yang$^{5}$\quad
  Chuanhao Li$^{3}$\\
  Xiaojie Xu$^{2}$\quad
  Hui He$^{4}$\quad
  Kaipeng Zhang$^{2}$~\footnotemark[2]\\[10pt]
  $^1$Tsinghua University\quad
  $^2$Shanda AI Research Tokyo\quad
  $^3$Shanghai AI Lab\quad \\
  $^4$Harbin Institute of Technology\quad
  $^5$Peking University\\
  \texttt{gin2pws@gmail.com \quad kaipeng.zhang@shanda.com}\\
}

\begin{document}
\maketitle
\begingroup
\renewcommand{\thefootnote}{\fnsymbol{footnote}}
\setcounter{footnote}{0}
\footnotetext[1]{Equal contribution.}
\footnotetext[2]{Corresponding author.}
\footnotetext[3]{Project leader.}
\endgroup
\begin{abstract}
Recent text-to-video generation models have made remarkable progress in visual realism, motion fidelity, and text-video alignment, yet they still struggle to produce socially coherent behavior. Unlike humans, who readily infer intentions, beliefs, emotions, and social norms from brief visual cues, current models often generate literal scenes without capturing the underlying causal and psychological dynamics. To systematically assess this limitation, we introduce the first benchmark for social reasoning in video generation. Grounded in developmental and social psychology, the benchmark covers thirty classic social cognition paradigms spanning seven core dimensions: mental-state inference, goal-directed action, joint attention, social coordination, prosocial behavior, social norms, and multi-agent strategy. To operationalize these paradigms, we build a fully training-free agent-based pipeline that distills the reasoning structure of each paradigm, synthesizes diverse video-ready scenarios, enforces conceptual neutrality and difficulty control through cue-based critique, and evaluates generated videos with a high-capacity VLM judge along five interpretable dimensions of social reasoning. Using this framework, we conduct the first large-scale evaluation of seven state-of-the-art video generation systems. Results show a clear gap between surface-level plausibility and deeper social reasoning, suggesting that current models remain limited in their ability to generate socially grounded behavior. \url{https://github.com/Gloria2tt/SVBench-Evaluation}
\end{abstract}    
\section{Introduction}

\label{sec:intro}

\begin{figure}[t]
  \centering
  \begin{subfigure}{\linewidth}
    \centering
    \includegraphics[width=1.0\linewidth]{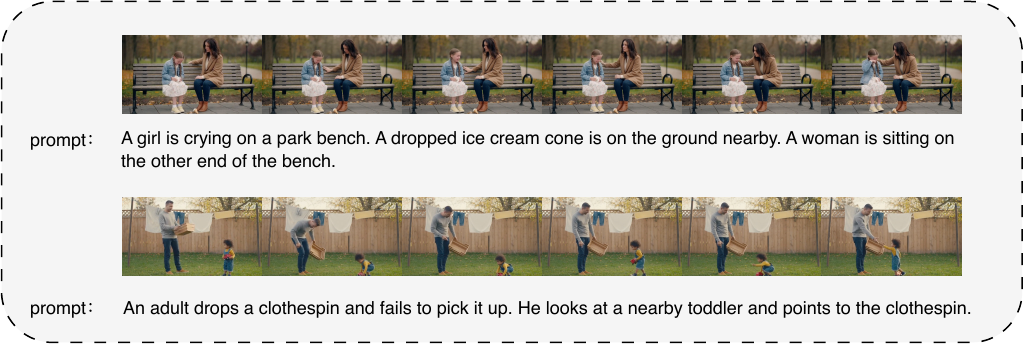}
    \caption{}
    \label{fig:mot1}
  \end{subfigure}

  \vspace{0.4em}

  \begin{subfigure}{\linewidth}
    \centering
    \includegraphics[width=1.0\linewidth]{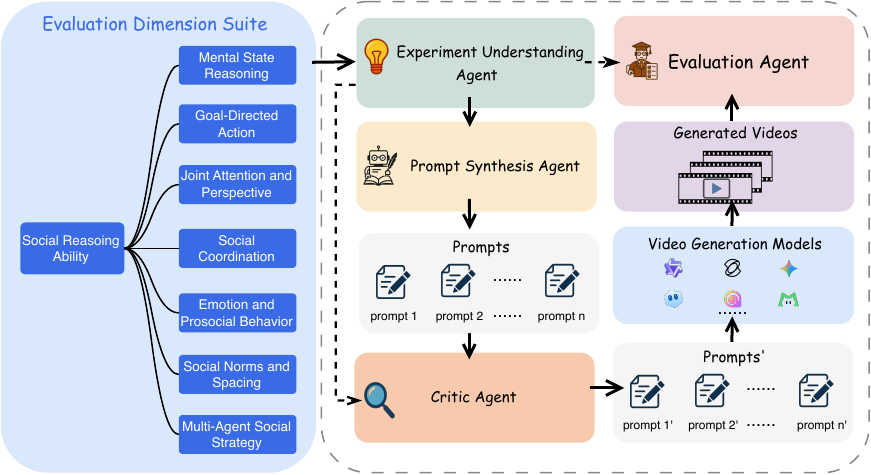}
    \caption{}
    \label{fig:mot2}
  \end{subfigure}

  \caption{\textbf{(a) Social reasoning scenario. 
(b) Our benchmark framework, which uses a two-part agent-based pipeline for constructing and evaluating social reasoning tasks in video generation.}}
  \label{fig:mot}
\end{figure}

Recent advances in video generation~\cite{kling,sora2,hailuo} have dramatically improved visual fidelity, temporal consistency, and text–video alignment. Modern diffusion- and transformer-based architectures can now synthesize dynamic scenes with striking realism, reproducing nuanced lighting, complex motion patterns, and multi-agent interactions across a wide range of environments. Yet, despite these impressive perceptual capabilities, today’s video generation models remain fundamentally socially unaware. They can reproduce what happens in the physical world, but struggle to represent why people act the way they do—failing to capture the latent beliefs, intentions, emotions, and norms that structure real human interactions.

From a cognitive perspective, this limitation is consequential. Decades of developmental psychology show that humans rely on deeply structured social reasoning to interpret visual scenes: whether in theory-of-mind tasks~\cite{premack1978does, baron1985does}, goal-inference experiments~\cite{woodward1998infants}, or studies of helping and joint attention~\cite{carpenter1998social}, people routinely make high-level inferences about others’ mental states and social motivations. These abilities emerge early in childhood and form the foundation for understanding everyday human activity. In contrast, existing video-generation benchmarks focus almost exclusively on low-level perceptual or physical factors—such as motion smoothness, visual quality, or physics plausibility—as seen in VBench~\cite{huang2024vbench}, EvalCrafter~\cite{liu2024evalcrafter}, T2V-CompBench~\cite{sun2025t2v}, and Morpheus~\cite{zhang2025morpheus}. While valuable, these metrics primarily assess whether a model can produce visually or physically plausible events, rather than whether it can generate socially appropriate, causally interpretable behavior when the target outcome is not explicitly specified in the prompt.

To illustrate this critical gap, consider two representative scenarios. First, imagine a crying girl sitting on a park bench next to a fallen ice cream cone, with a woman nearby (Fig.~\ref{fig:mot1}). Humans instantly infer the causal link between the dropped ice cream and the girl’s distress and naturally anticipate that the adult may comfort her—an effortless chain of intention, emotion, and theory-of-mind reasoning. In another scenario, an adult drops a clothespin, fails to retrieve it, and then points toward it while looking at a nearby toddler. Humans readily interpret this gesture as a request for help, expecting that the child might attempt instrumental helping; in fact, developmental studies (e.g.,~\cite{meltzoff1995understanding}) show that even 14--18-month-old infants understand such unfulfilled goals and cues. But when these same descriptions are provided as input to a video generation model, will the resulting videos exhibit these socially meaningful inferences? Will the model generate comforting behavior, causal linking, or helping actions—or will it merely render a literal visual scene without the underlying social logic? These examples highlight a fundamental distinction: physical reasoning determines how events unfold visually, whereas social reasoning governs whether agents behave in ways that are socially and causally appropriate. Current video generation systems excel at the former but remain limited on the latter.

In this paper, we directly target this gap by introducing a benchmark for \emph{social reasoning in video generation}. As illustrated in Figure~\ref{fig:mot2}, our framework is anchored in well-established findings from developmental and social psychology, which converge on seven core components of social cognition: mental-state inference~\cite{premack1978does, baron1985does}, goal-directed action understanding~\cite{csibra2008goal}, joint attention and perspective-taking, social coordination~\cite{tomasello2005understanding}, emotion and prosocial responding~\cite{warneken2006altruistic}, social norms and interpersonal spacing~\cite{hall1966hidden}, and multi-agent social strategy~\cite{whiten1988machiavellian}. These capacities naturally map onto the dynamic, causally structured nature of video generation, where models must not only render realistic frames but also produce behavior that unfolds coherently over time.

To operationalize this taxonomy, we selected thirty classic psychological experiments that collectively span these seven dimensions. We then developed a fully training-free, agent-based pipeline to construct and evaluate video-generation tasks. This pipeline comprises four components:
(1) an Experiment Understanding Agent, which processes the description of each psychological experiment and distills its underlying social reasoning mechanism. This step ensures that downstream generation is grounded in the intended cognitive construct rather than superficial scenario details;
(2) a Prompt Synthesis Agent, which elaborates each experiment into multiple concrete scenarios by varying agent identities, object layouts, and environmental contexts. This agent translates abstract cognitive paradigms into visually grounded, video-ready situations, enabling systematic generalization across diverse settings;
(3) a Critic Agent, which performs two critical functions: (a) enforcing conceptual neutrality by removing descriptive elements that might reveal the “correct answer” and thus compromise evaluation integrity; and (b) generating difficulty-controlled variants by manipulating social cues—such as gaze direction, occlusion, or affordance visibility—to produce easy, medium, and hard versions that probe the robustness of a model’s social reasoning;
(4) an Evaluation Agent (EVA), implemented using a high-capacity vision–language model, which assesses generated videos along five discrete, interpretable dimensions of social reasoning quality. EVA first reconstructs the expected logic of the experiment, then evaluates whether the video exhibits appropriate causal structure, social cues, and behavioral plausibility.

Together, this agent-based pipeline offers a scalable, controlled, and theoretically grounded framework for constructing and evaluating social reasoning tasks in video generation models, enabling large-scale assessment without reliance on human annotation.

Our main contributions can be discribe as follows:
\begin{itemize}
    \item We introduce the first benchmark specifically designed to evaluate social reasoning in video generation, grounded in seven core capacities identified in developmental and social psychology.
    \item We design a training-free, four-agent pipeline capable of automatically constructing difficulty-controlled scenarios and evaluating model outputs at scale.
    \item Through extensive experiments across eight state-of-the-art video generators, we provide the first systematic analysis revealing where current models succeed or fundamentally fail in generating socially coherent behavior.
\end{itemize}

\section{Related Work}
\subsection{Evaluation Benchmarks for Video Generation}
Multimodal technologies~\cite{peng2024t3m,peng2024data,mao2025yume} have progressed rapidly in recent years, laying the foundation for modeling complex signals. 
Early evaluations of video generation models focused on perceptual fidelity and temporal stability, adapting video quality metrics FVD~\cite{unterthiner2019fvd} and human preference scores\cite{liu2025improving}. Recent benchmarks have shifted toward interpretable, axis-wise diagnostics that decompose generation quality into orthogonal dimensions. VBench~\cite{huang2024vbench, huang2025vbench++} evaluates models across fine-grained aspects such as subject consistency, background consistency, temporal flickering, motion smoothness, and aesthetic quality. EvalCrafter consolidates metrics for visual quality, content consistency, motion realism, and text-video alignment. VBench-2.0~\cite{zheng2025vbench} extends evaluation to higher-order axes such as human fidelity, controllability, creativity, physics plausibility, and commonsense reasoning. Physics-centered suites explicitly test adherence to physical laws: PhyCoBench~\cite{chen2025physical} measures physical~\emph{coherence} with optical-flow–guided frame prediction and automated scoring aligned to human assessment; Morpheus uses real physical experiments and conservation-law–based probes to benchmark \emph{physical reasoning} in generated videos. While these benchmarks push evaluation beyond appearance, they still focus primarily on perceptual quality, physical plausibility, or action-level consistency. By contrast, \emph{social reasoning}---involving multi-agent interactions, roles, norms, intentions, and commonsense social dynamics---remains under-explored. Our work fills this gap by designing a benchmark that evaluates whether models generate socially appropriate, paradigm-consistent behavior when the intended outcome is not explicitly specified in the prompt, rather than measuring low-level action or physical plausibility.

\subsection{Social Reasoning in AI System}
Social reasoning---the ability to attribute mental states, infer intentions, and act in accordance with social norms---has been recognized as an important ingredient of more human-like AI. In large language models (LLMs), benchmarks such as~\cite{kosinski2023theory, ullman2023large, xu2024opentom} probe Theory of Mind (ToM) and multi-agent belief tracking. These studies show that while current models can handle simple first-order belief narratives, they remain unreliable in higher-order or counterfactual settings where an agent must reason about \emph{what another agent believes}. In the video domain, social intelligence has largely been studied as an \emph{analysis} problem. Social-IQ \cite{zadeh2019social} evaluates emotion recognition and social situation understanding from human-created video clips, while recent VideoQA benchmarks such as R\textsuperscript{3}-VQA \cite{niu2025r} introduce fine-grained annotations of social events, mental states, and causal social chains to assess social reasoning through question-answering tasks. However, these approaches fundamentally evaluate whether models can \emph{interpret} social cues in existing videos—they do not assess whether a model can \emph{generate} socially coherent multi-agent interactions from scratch. This represents a critical gap: while discriminative benchmarks test whether models recognize social reasoning in human-created content, no prior work evaluates whether video generation models can \emph{synthesize} it—whether they can produce scenarios where agents exhibit plausible beliefs, respond to others' mental states, and behave according to social norms. Our benchmark addresses this gap by introducing the first systematic evaluation of social reasoning capabilities in generative video models.

\section{Method}

\begin{figure*}[t]
    \centering
    \includegraphics[width=0.9\textwidth]{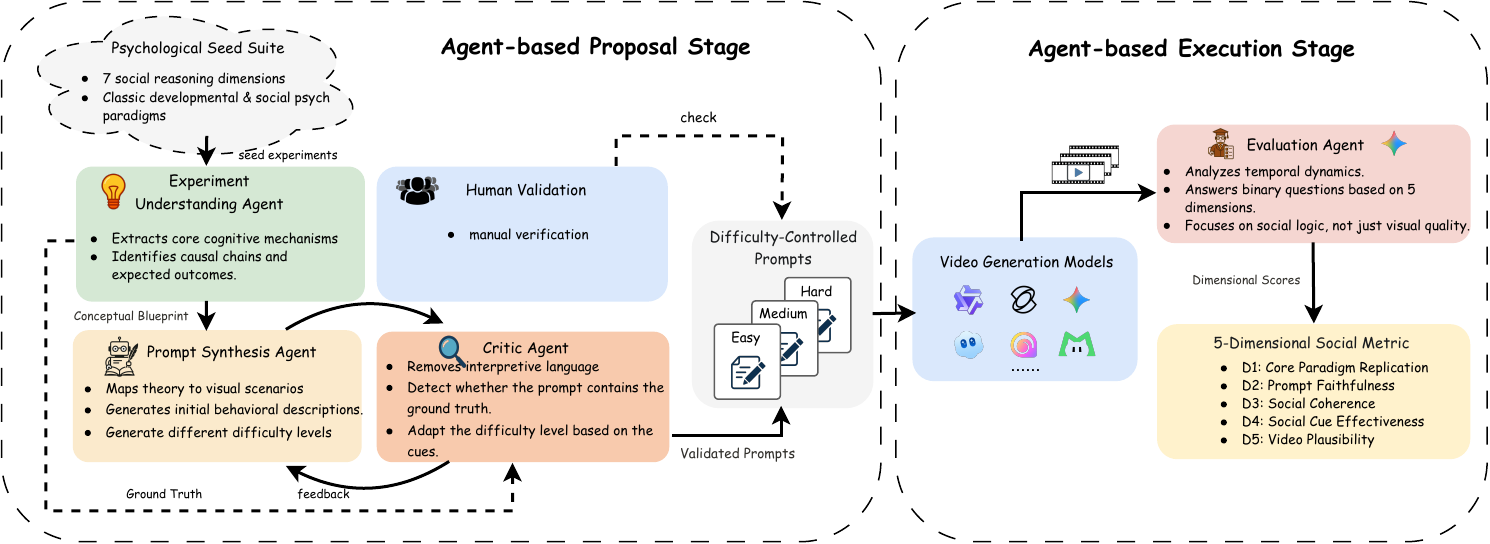}
    \caption{Pipeline overview. The framework consists of two training-free components: 
    (1) an \emph{agent-based generation pipeline} that transforms psychologically grounded social reasoning experiments into diverse, difficulty-controlled video prompts, 
    and 
    (2) an \emph{agent-based evaluation pipeline} that uses a vision–language model to score generated videos along five discrete dimensions of social reasoning. }
    \label{fig:pipeline}
\end{figure*}

\subsection{Seed Suite and Feasible Subset}
\label{sec:seed-filter}

We construct a seed suite of thirty social reasoning experiments grounded in developmental and social psychology, spanning seven dimensions of social cognition: Mental State Reasoning, Goal–Directed Action, Joint Attention and Perspective, Social Coordination, Emotion and Prosocial Behavior, Social Norms and Spacing, and Multi–Agent Social Strategy. Each dimension corresponds to established experimental paradigms, providing strong theoretical grounding and interpretability.

Considering current video generation systems can typically produce only short clips (5–10 seconds) with one or two salient actions, we partition these thirty experiments into two groups. The first group consists of tasks whose core social reasoning cues can be fully expressed within a short video—requiring only a single scene, a small number of agents, and visually explicit cues such as gaze, gesture, posture, or spatial configuration. These constitute the fifteen \emph{short-video–feasible} experiments used as the primary benchmark in this paper. The second group comprises tasks whose reasoning structure unfolds over multiple events or extended temporal sequences (e.g., delayed gratification, multi-step deception, multi-stage joint planning), making them unsuitable for today’s short-form video generators. These constitute the fifteen \emph{long-horizon} experiments, which we include as an extended benchmark appendix for future long-video generation models.

The distribution of short-video–feasible and long-horizon tasks across the seven social reasoning dimensions is summarized in Table~\ref{tab:chosen_deleted_7dims}.

\subsection{Agent-Based Generation}
Our agent-based generation framework comprises three agents: the Experiment Understanding Agent, the Prompt Synthesis Agent, and the Critic Agent. Figure~\ref{fig:pipeline} depicts the pipeline, and the following sections describe each component in detail.

\paragraph{Seed Experiment Understanding} As shown in Figure~\ref{fig:pipeline}, in the initial stage, we employ the Experiment Understanding Agent to create comprehensive conceptual analyses for each seed experiment. Specifically, the agent generates a structured understanding containing four essential components: a detailed description that formalizes the psychological phenomenon being tested, key concepts identifying the relevant cognitive mechanisms, a test point articulating the specific reasoning capability under evaluation, and ground truth specifying the expected behavioral outcome or correct interpretation. This structured specification ensures that subsequent prompt generation maintains theoretical fidelity to established psychological research while being tailored for video generation evaluation. The explicit decomposition forces the model to reason about experimental design before generating concrete scenarios, reducing conceptual drift and creating an interpretable intermediate representation.

\begin{table*}[t]
\scriptsize
\setlength{\tabcolsep}{3pt}
\renewcommand{\arraystretch}{1.0}
\centering
\caption{Overview of the thirty seed experiments and their categorization into short-video–feasible tasks (selected) and long-horizon tasks (excluded), grouped by seven dimensions of social reasoning.}
\label{tab:chosen_deleted_7dims}
\begin{tabularx}{\textwidth}{l *{7}{>{\raggedright\arraybackslash}X}}
\toprule
& \textbf{Mental State Reasoning} & \textbf{Goal Directed Action} & \textbf{Joint Attention and Perspective} & \textbf{Social Coordination} & \textbf{Emotion and Prosocial Behavior} & \textbf{Social Norms and Spacing} & \textbf{Multi Agent Social Strategy} \\
\midrule
\textbf{short} &
--- &
Detour Reaching \cite{lockman1984development}; Tool Selection \cite{defeyter2009developmental} &
Gaze Following \cite{friesen1998eyes}; Pointing Comprehension \cite{tomasello2007new}; Joint Engagement \cite{bakeman1984coordinating} &
Turn Taking \cite{duncan1972some}; Leader Follower Coordination \cite{marsh2009social} &
Emotion Contagion \cite{hatfield1993emotional}; Instrumental Helping \cite{warneken2006altruistic}; Empathic Concern \cite{eisenberg1990empathy}; Costly Helping \cite{batson2014altruism} &
Proxemics Personal Space \cite{hall1966hidden}; Queue Behavior \cite{mann1970social}; Dominance Display Posture Space \cite{carney2005beliefs} &
Helping Based on Visual Perspective \cite{moll200714} \\
\midrule
\textbf{long} &
Sally-Anne Test \cite{baron1985does}; Smarties Task \cite{perner1987three}; Level 2 Visual Perspective Taking \cite{flavell1977development}; Knowledge Access \cite{pratt1990young}; Intentional vs Accidental Actions \cite{malle1997folk} &
Kohler Stick \cite{kohler2018mentality}; Gergely's Head-Touch \cite{csibra2008goal}; Failed Attempts \cite{meltzoff1995understanding} &
Level 1 Visual Perspective Taking \cite{masangkay1974early} &
Collaborative Transport \cite{warneken2006cooperative}; Collision Avoidance Pedestrian Flow \cite{helbing1995social} &
--- &
--- &
Competitive Resource Allocation \cite{fehr1999theory}; Cooperative Deception Detection \cite{whiten1988machiavellian}; Norm Violation Response \cite{rakoczy2008taking}; Multi Party Collaborative Problem Solving Asymmetric Information \cite{tomasello2005understanding} \\
\bottomrule
\end{tabularx}
\vspace{-0.6em}
\end{table*}

\paragraph{Prompt Synthesis}
Building upon the experiment specifications (see Figure~\ref{fig:pipeline}), the Prompt Synthesis Agent operationalizes abstract social reasoning concepts into observable action sequences. The generation process adheres to four key principles designed specifically for video generation evaluation:

\begin{enumerate}
\item \textbf{Action-Oriented Description:} Prompts describe exclusively visually observable behaviors, deliberately excluding internal mental states or expected outcomes to prevent ``teaching to the test.''
\item \textbf{Temporal Feasibility:} All scenarios are designed for 5--10 second video clips, aligning with current model capabilities.
\item \textbf{Concrete Instantiation:} Abstract social concepts are embodied through specific entities (e.g., particular age groups, genders, species) rather than abstract placeholders.
\item \textbf{Evaluation Readiness:} Each prompt maintains clear separation between action descriptions and expected outcomes, ensuring unbiased assessment.
\end{enumerate}
Together, these principles ensure that generated prompts are both psychologically meaningful and practically viable for benchmarking video generation systems on nuanced social reasoning tasks. 

\paragraph{Critic Agent}
The Critic Agent examines each synthesized prompt and enforces three requirements: (1) removal of interpretive language, (2) detection of ground-truth leakage, and (3) difficulty control through cue manipulation.

First, it eliminates \emph{mental-state or interpretive phrasing} such as ``realizes", ``feels sad", or ``decides to help", ensuring that prompts describe only observable behavior. For example, a sentence like ``The woman realizes the man cannot reach the book and decides to help" is rewritten as “The man stretches toward a book on a high shelf but cannot reach it; the woman notices this and walks toward the shelf", keeping all cues strictly behavioral. Second, the critic checks for \emph{ground-truth leakage} by comparing the prompt with the experiment’s test point. If the correct outcome is explicitly stated—e.g., describing that the bystander helps, selects the right tool, or finds the hidden object—the critic flags the violation and returns structured correction instructions to the Prompt Synthesis Agent. Third, the critic regulates \emph{difficulty} by adjusting the presence of psychological (gaze, facial expression), motoric (reaching, pointing), and contextual (object placement, affordance) cues. Easy variants include redundant cues, medium variants retain only those minimally required for inference, and hard variants remove or obscure central cues to require more subtle reasoning.

Whenever a prompt fails any of these checks, the Critic Agent does not simply reject it; instead, it returns explicit diagnostic feedback (violation type and suggested edits) to the Prompt Synthesis Agent. The generation module then regenerates or revises the prompt under these additional constraints. This iterative loop continues until the prompt satisfies neutrality, no-leakage, and difficulty requirements, yielding a pool of validated, difficulty-controlled prompts for each experiment.

\subsection{Agent-Based Evaluation}

Evaluating social reasoning in generated videos poses a unique challenge, as social interactions lack a single canonical ground truth, unlike tasks with deterministic outcomes. A prompt describing a helping scenario, for example, can be realized through countless valid behaviors. Consequently, our evaluation must shift focus from fidelity to a specific reference video towards assessing whether the intended \emph{social logic} of the experimental paradigm emerges correctly.

To this end, we introduce an agent-based evaluation framework that utilizes a Vision-Language Model (VLM) as a structured judge. We deliberately avoid continuous scores, which suffer from significant noise and instability due to the VLM's difficulty in calibrating a fine-grained numerical scale across diverse prompts. Instead, we propose five binary evaluation dimensions. This discrete approach enhances robustness by framing the evaluation as a series of unambiguous factual questions (e.g., "Did the agent react based on what it could see?"). This aligns more closely with human categorical judgments and substantially reduces inter-trial variance in VLM assessments.

For each generated video, the VLM judge is provided with minimal experimental metadata and evaluates the output along five dimensions. \textbf{D1: Core Paradigm Replication} assesses if the core psychological phenomenon is correctly instantiated. \textbf{D2: Prompt Faithfulness} ensures adherence to the specified agents, objects, and scene, preventing semantic circumvention. \textbf{D3: Social Coherence} checks for causally and socially plausible agent behaviors. \textbf{D4: Social Cue Effectiveness} evaluates the rendering of critical perceptual cues like gaze and gestures. Finally, \textbf{D5: Video Plausibility} serves as a baseline for visual stability, isolating generation failures from reasoning errors. Each dimension $D_k$ is scored as $\{0,1\}$, and the overall score is the average of these assessments:
\[
S_{\mathrm{overall}} = \frac{1}{5} \sum_{k=1}^{5} D_k.
\]
This structured design yields three key advantages: it enables \textbf{disentanglement} of failure modes (e.g., generation vs. reasoning), ensures \textbf{robustness} by minimizing VLM calibration noise, and provides \textbf{scalability} for future extension to more complex social scenarios.

\section{Experiment}
Given that contemporary video generation models are typically limited to producing short clips of 5--10 seconds, our main experiments focus on the 15 \emph{short-form} social reasoning tasks identified in Section~\ref{sec:seed-filter}. For each task, we design three difficulty levels and three prompts per level, resulting in a total of 135 evaluation prompts. These tasks can be fully expressed within a single short video segment, making them suitable for evaluation with current-generation models; results on the long-form tasks are provided in the supplementary material.

\begin{table*}[t]
\centering
\scriptsize
\caption{\textbf{Experimental Results of Selected 15 Tasks Across 8 Models.} We report the performance (\%) of eight models on tasks grouped by social reasoning dimensions.}
\label{tab:experiment_results}
\setlength{\tabcolsep}{4pt}
\begin{tabular}{l@{\hskip 1em}l@{\hskip 1em}cccccccc}
\toprule
\multirow{2}{*}{\textbf{Task Dimension}} & \multirow{2}{*}{\textbf{Sub-Task}} & \multicolumn{8}{c}{\textbf{Model Performance (\%)}} \\
\cmidrule(lr){3-10}
 & & 
 \rotatebox{45}{\textbf{Hailuo02-S}} & 
 \rotatebox{45}{\textbf{kling2.5-turbo}} & 
 \rotatebox{45}{\textbf{Sora2pro}} & 
 \rotatebox{45}{\textbf{Veo-3.1}} & 
  
 \rotatebox{45}{\textbf{HunyuanVideo}} & 
 \rotatebox{45}{\textbf{Longcat-Video}} & 
 \rotatebox{45}{\textbf{LTX-1.0}} &
 \rotatebox{45}{\textbf{Wan2.2}} \\
\midrule    
\multirow{2}{*}{Goal Directed Action} 
 & Detour Reaching & 51.4 & 48.6 & \textbf{68.6} & 62.9 &  31.4 & 28.6 & 17.1 & 42.2\\
 & Tool Selection & 55.0 & 45.0 & \textbf{85.0} & \textbf{85.0} & 17.5 & 28.6 & 30.0 & 57.8\\
\midrule
\multirow{3}{*}{\shortstack[l]{Joint Attention \&\\Perspective }}
 & Gaze Following & 44.4 & 33.3 & 62.2 &  \textbf{68.9} & 35.6 & 44.4 & 28.9 & 28.9 \\
 & Pointing Comprehension & 75.0 & 67.5 & \textbf{87.5} & 82.5 & 30.0 & 57.1 & 22.9 &71.1\\ 
 & Joint Engagement & 45.0 & 45.0 & 82.5 & 82.5 & 50.0 & 31.4 & 30.0 & 44.4 \\
\midrule      
\multirow{2}{*}{Social Coordination }
 & Turn Taking & 45.7 & 62.9 & \textbf{94.3} & 85.7 &  37.1 & 57.1 & 40.0 & 62.2\\
 & Leader Follower Coord & 40.0 & 44.4 & \textbf{77.8} & 64.4 & 17.8 & 35.6 & 17.8 & 48.9\\ 
\midrule
\multirow{4}{*}{\shortstack[l]{Emotion \& Prosocial\\Behavior }}
 & Emotion Contagion & 66.7 & 75.6 & 82.2 & \textbf{88.9} &  46.7 & 65.0 & 57.8 & \textbf{88.9}\\  
 & Instrumental Helping & \textbf{68.9} & 55.6 & 62.2 & 48.9 &  31.1 & 33.3 & 24.4 & 35.6\\     
 & Empathic Concern & 80.0 & 76.0 & \textbf{100.0} & \textbf{100.0} &  24.0 & 60.0 & 20.0 & 66.7 \\        
 & Costly Helping & 57.5 & 37.5 & \textbf{95.0} & 75.0 &  22.5 & 31.4 & 15.0 & 37.8\\
\midrule 
\multirow{3}{*}{\shortstack[l]{Social Norms \&\\Spacing}}
 & Proxemics Personal Space & 47.5 & 47.5 & \textbf{65.0} & 45.0 &  25.0 & 25.0 & 35.0 & 22.2\\  
 & Queue Behavior & 55.6 & 40.0 & \textbf{82.5} & 75.6 &  33.3 & 20.0 & 20.0 & 20.0\\
 & Dominance Display & 80.0 & 75.0 & 75.0 & \textbf{82.5} & 35.0 & 30.0 & 37.5 & 64.4\\   
 
Multi-Agent Social Strategy 
 & Helping Based on Visual Perspective & 42.2 & 42.2 & \textbf{84.4} & 53.3 &  24.4 & 40.0 & 17.8 & 33.3\\
 \midrule
 Overall 
 & - & 56.4 & 52.2 & \textbf{79.6} & 72.4 & 30.8 & 39.2 & 27.6 & 48.3\\
\bottomrule
\end{tabular}
\end{table*}

\subsection{Experiment Setup}
We evaluate our benchmark on a diverse set of state-of-the-art text-to-video generation systems, covering both proprietary and commercial-grade models. Concretely, for closed-source model we include four representative models: Sora2pro~\cite{sora2}, Kling2.5turbo~\cite{kling}, Veo3.1~\cite{veo3}, Hailuo02 Standard(Hailuo02-S)~\cite{hailuo}. For open-source model, we select three representative models: Hunyuan-Video~\cite{kong2024hunyuanvideo}, LTX-1.0~\cite{hacohen2024ltx}, LongCat-Video~\cite{team2025longcat} and Wan2.2~\cite{wan2025wan}. We present our experiments in the following sections. For evaluation, we use Gemini 2.5 Pro~\cite{comanici2025gemini} as our vision–language model (VLM) judge.

\subsection{Evaluation Results}
Table \ref{tab:experiment_results} summarizes the performance of representative text-to-video models across the 15 socially grounded tasks, grouped into six major dimensions of social reasoning. 

First, Sora2-Pro and Veo-3.1 exhibit a clear advantage across nearly all task categories, achieving overall pass rates of 79.6\% and 72.4\%, respectively. Their strengths are particularly pronounced in tasks involving goal understanding, joint attention, and prosocial behavior, where both models exceed 80\% on most sub-tasks. These results suggest that top-tier proprietary systems already possess strong implicit priors for human motion causality, gaze direction, and intention-driven interactions, even without explicit cue engineering.
In contrast, Hailuo02-S and Kling2.5-Turbo show substantially weaker reasoning ability, with overall scores of 56.4\% and 52.2\%. These models struggle especially with tasks that require coordinated multi-agent behavior (e.g., Leader–Follower Coordination) or abstract social inference (e.g., Helping Based on Visual Perspective), exhibiting failure rates over 50\%. Their performance improves noticeably when explicit cues are available (e.g., Pointing Comprehension), indicating a higher reliance on surface-level visual signals.

A significant performance gap also separates proprietary systems from open-source models. Longcat-Video, HunyuanVideo, and LTX-1.0 operate at a substantially lower level across nearly all dimensions, with particularly poor performance on tasks requiring complex causal or belief-state reasoning. Wan2.2 is the strongest open-source model, showing particular strength on \textit{Emotion Contagion} and \textit{Pointing Comprehension}. Even so, it still trails the best proprietary systems, underscoring the remaining gap in sophisticated social reasoning.



\subsection{Verification of the Agent-Based Generation}

\paragraph{Validation of Prompt Quality Across Pipeline Stages.}

\begin{table}[t]
\centering
\small  
\setlength{\tabcolsep}{3.5pt}  
\caption{Human pass rates (\%) for prompts from pipeline stages:
(1) No Understanding, (2) +Synthesis, (3) Full pipeline.}
\begin{tabular}{lccc}
\toprule
\textbf{Dimension} & \textbf{No Und.} & \textbf{+Synth.} & \textbf{Full} \\
\midrule
Goal Directed Action & 68.1 & 76.5 & 87.5 \\
Joint Attention \& Perspective & 66.5 & 75.2 & 86.3 \\
Social Coordination & 67.2 & 74.5 & 86.5 \\
Emotion \& Prosocial & 68.3 & 78.3 & 88.2 \\
Social Norms \& Spacing & 66.4 & 77.2 & 87.2 \\
Multi-Agent Strategy & 65.6 & 73.5 & 85.6 \\
Mental State Reasoning & 65.8 & 76.1 & 87.2 \\
\midrule
\textbf{Average} & \textbf{66.8} & \textbf{75.9} & \textbf{86.9} \\
\bottomrule
\end{tabular}
\label{tab:3}
\end{table}
A key contribution of our benchmark is the agent-based pipeline that transforms abstract psychological paradigms into concrete, video-ready prompts. To validate the quality of the constructed prompts, we conduct a human evaluation comparing three stages of the generation process.

 We test prompts generated at three stages of the pipeline:
(1) without any conceptual understanding (“No Understanding”),
(2) with Experiment Understanding followed by Prompt Synthesis (“Understanding + Synthesis”), and
(3) the full pipeline including Critic Agent revision (“Full”).
Human judges simply decide whether a prompt correctly expresses the intended reasoning construct while remaining descriptively neutral and free of answer leakage.

As shown in Table~\ref{tab:3}, incorporating conceptual understanding substantially improves prompt validity across all seven social reasoning dimensions. Pass rates rise from roughly 67.7\% without understanding to over 75.8\% once the test point is explicitly distilled, and further to above 86.7\% when Critic Agent refinement is applied. These results demonstrate that the reasoning-aware generation stage—and its subsequent critic-driven correction—are both essential for producing prompts that are theoretically faithful and suitable for evaluating social reasoning in video generation models.

\paragraph{Cue-Controllability and Difficulty Ordering.}
\begin{table}[t]
\caption{Average pass rates (\%) under different difficulty levels for four closed-source models. The results reflect how cue-based difficulty modulation affects each model’s performance.}
  \centering
  \begin{tabular}{@{}lccc@{}}
    \toprule
    Models  & Easy & Mid & Hard \\
    \midrule
    Sora2pro &73.8& 84.8&  79.4 \\
    Veo3.1  &66.6&  74.4& 75.8 \\
    Hailuo02-S  &62.6&  56.8&  49.8 \\
    Kling2.5turbo &58.0&  54.0& 44.6 \\
    \bottomrule
  \end{tabular}
  \label{tab:4}
\end{table}

We further validated that the easy, medium, and hard variants produced by the Critic Agent form a meaningful difficulty hierarchy that modulates the reasoning challenge faced by different video generation systems. As shown in Table~\ref{tab:4}, the cue-based variants yield a clear difficulty ordering for \textit{Hailuo02-S} and \textit{Kling2.5-Turbo} (Easy $>$ Medium $>$ Hard), confirming that richer social cues effectively facilitate reasoning in models with weaker inference capacity. However, \textit{Sora2-Pro} and \textit{Veo3.1} exhibit a reversed trend: their performance peaks at the medium or hard conditions despite reduced visual cues. We attribute this to their stronger intrinsic social reasoning capability—these models can infer social intent and causal structure even under minimal information, and additional cues may introduce conflicting or redundant signals that cannot be perfectly realized within a 5–10s clip, leading to penalties on other dimensions such as \textit{Prompt Faithfulness} or \textit{Social Cue Use}. In contrast, the weaker models benefit directly from explicit cues, showing a monotonic improvement when the task becomes more guided. Together, these results confirm that our cue-based difficulty design not only controls the inferential complexity of each prompt but also reveals distinct reasoning regimes across models: high-performing systems demonstrate robustness under cue sparsity, whereas lower-capacity models rely heavily on external cue enrichment.

\subsection{Validation of the Agent-Based Evaluation}

Our benchmark relies on an automated VLM Judge to assess social coherence, task correctness, and causal consistency in generated videos. To ensure that this automatic evaluation is trustworthy, we perform several analyses examining its reliability and alignment with human reasoning.

\paragraph{Human Alignment}
\begin{figure}[t]
  \centering
  \begin{subfigure}{\linewidth}
    \centering
    \includegraphics[width=0.9\linewidth]{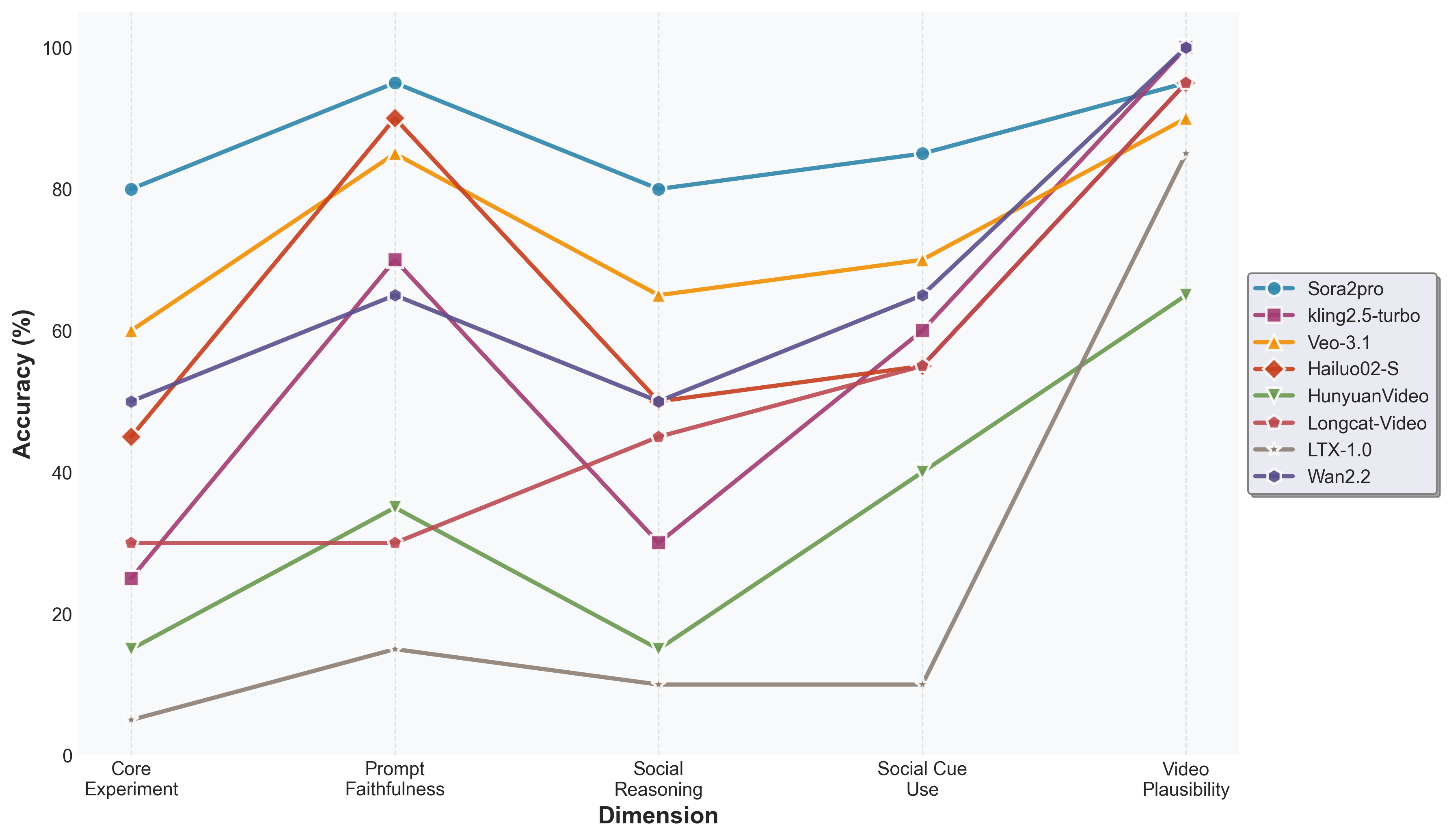}  
    \caption{Agent evaluation results}
    \label{fig:human_alignment_a}
  \end{subfigure}

  \vspace{0.4em}  

  \begin{subfigure}{\linewidth}
    \centering
    \includegraphics[width=0.9\linewidth]{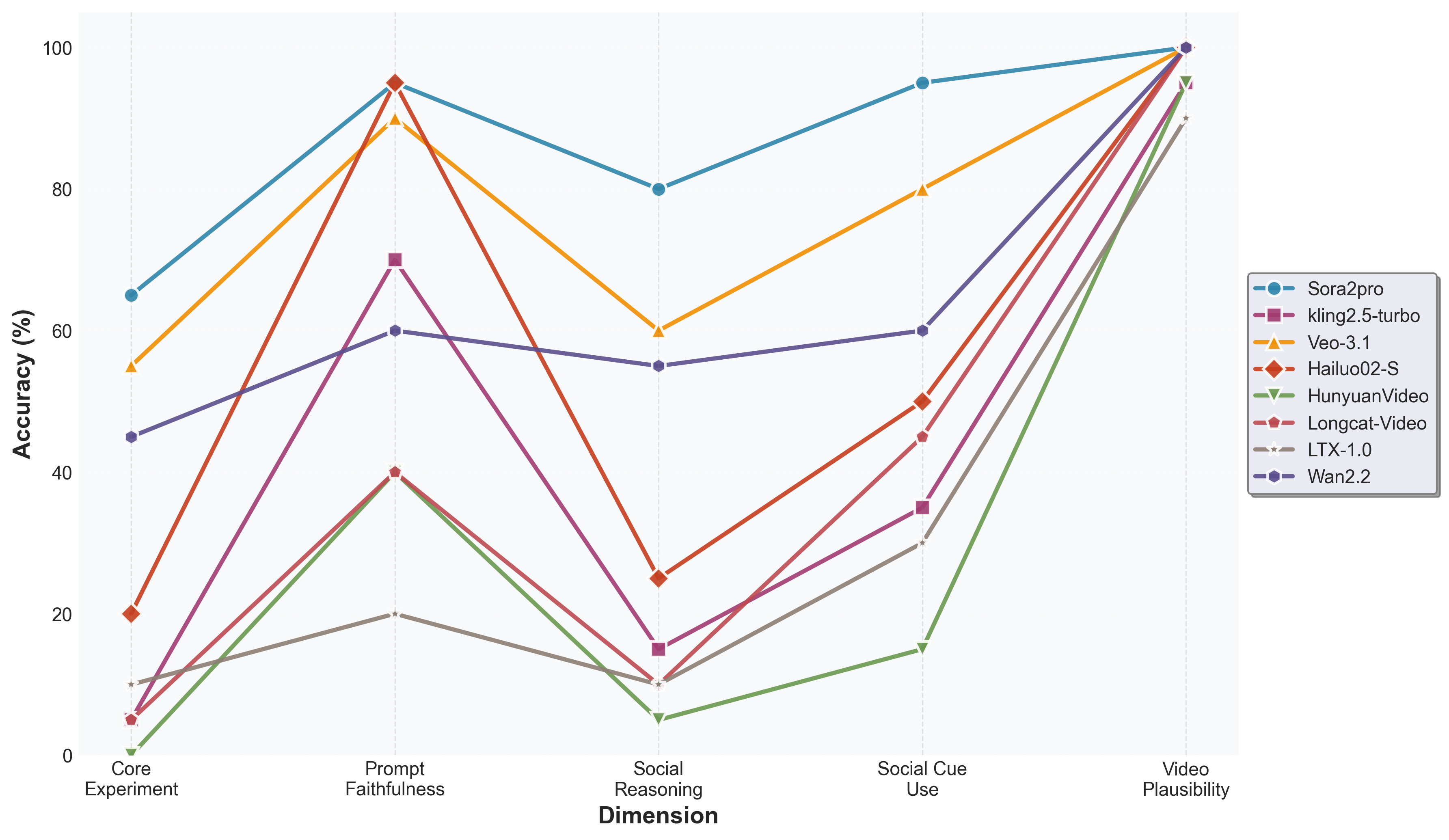}  
    \caption{Human evaluation results}
    \label{fig:human_alignment_b}
  \end{subfigure}

  \caption{Comparison between automated and human evaluation. (a) Scores from the agent-based VLM Judge. (b) Scores from human annotators using the same five-dimensional rubric. The two profiles exhibit closely matched trends across dimensions and models.}
  \label{fig:human_alignment}
\end{figure}
To validate our automated evaluation protocol, we utilize Gemini 2.5 Pro as our Vision-Language Model (VLM)-based judge. As exhaustive human annotation across all models is prohibitively expensive, we adopt a stratified sampling strategy and sample 20 generated videos per model, yielding a 160-clip test set. These clips are evaluated by 10 annotators, including project members and other graduate students, using the same five-dimensional rubric as the VLM judge, with multiple independent ratings per clip averaged to obtain the final human score. As shown in Figure~\ref{fig:human_alignment}, our comparative analysis reveals a high degree of alignment in the relative performance trends between the VLM judge (Fig.~\ref{fig:human_alignment_a}) and human annotators (Fig.~\ref{fig:human_alignment_b}). The relative difficulty of the evaluation dimensions remains consistent for both. However, we identify a systematic divergence in scoring thresholds. Human annotators are more lenient on perceptual dimensions like \emph{Prompt Faithfulness (D2)}, \emph{Social Cue Use (D4)}, and \emph{Video Plausibility (D5)}, where their pass rates approach ceiling. This suggests humans prioritize sufficient visual clarity for interpretability over pixel-perfect generation.
Conversely, on reasoning-intensive dimensions—\emph{Core Experiment (D1)} and \emph{Social Reasoning (D3)}—human raters are markedly stricter. They demand that the complete causal and logical structure of the psychological paradigm be correctly instantiated, penalizing even minor deviations from the intended social logic. This intuitive bias reflects a tolerance for surface-level imperfections but an intolerance for logical flaws.

\paragraph{Qualitative Case Analysis}
\begin{figure}[t]
  \centering
  \begin{subfigure}{\linewidth}
    \centering
    \includegraphics[width=1.0\linewidth]{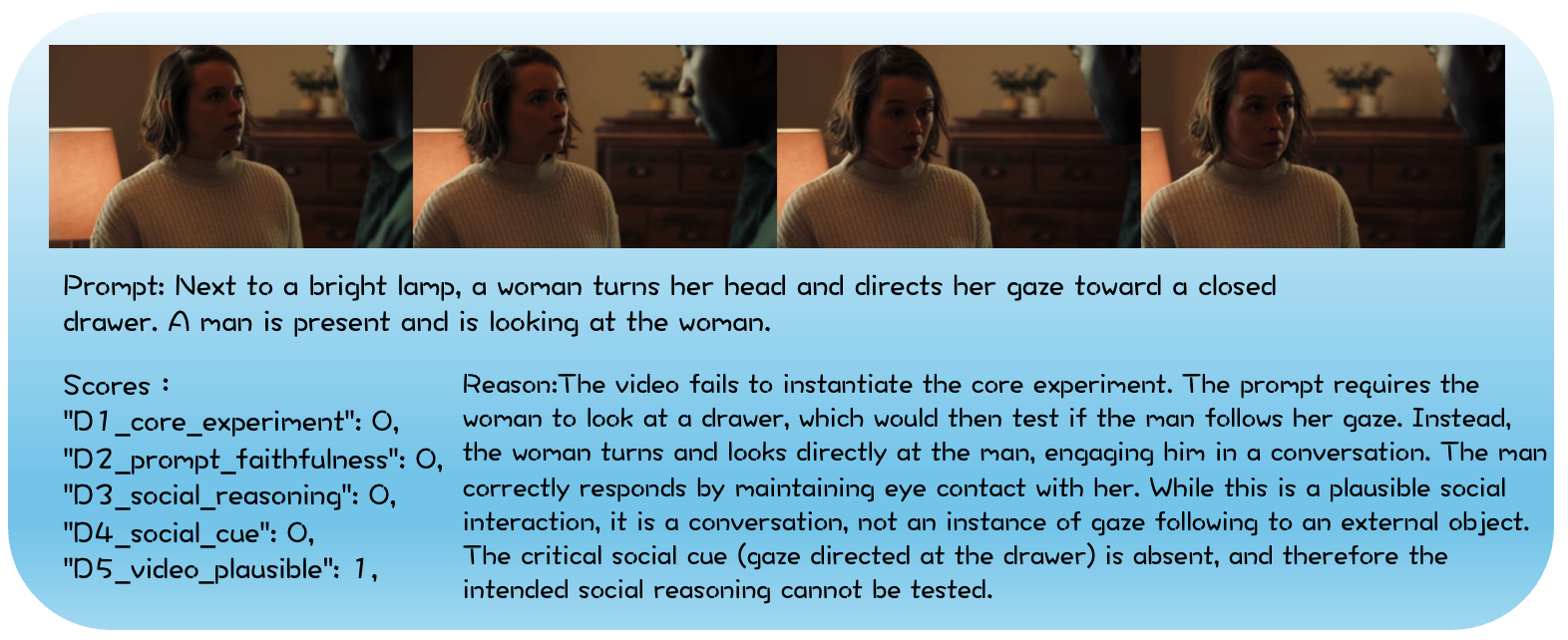}
    \caption{}
    \label{fig:visual1}
  \end{subfigure}

  \vspace{0.4em}

  \begin{subfigure}{\linewidth}
    \centering
    \includegraphics[width=1.0\linewidth]{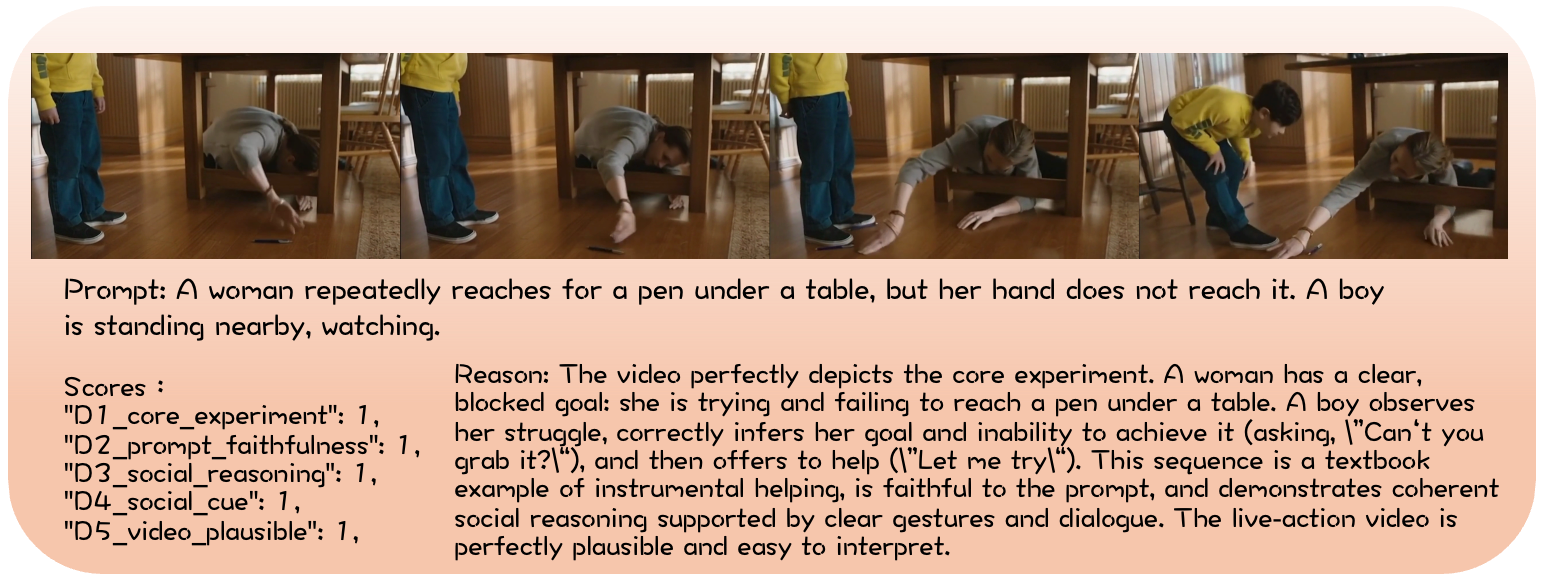}
    \caption{}
    \label{fig:visual2}
  \end{subfigure}

  \caption{Representative qualitative cases of the agent-based evaluator. Each panel shows sampled video frames, the generation prompt, the five-dimensional scores, and the model’s natural-language justification.}
  \label{fig:visualiz}
\end{figure}

To illustrate the practical application of our framework, Figure~\ref{fig:visualiz} presents two representative cases that showcase the VLM judge's decision-making process.

The first case (Fig.~\ref{fig:visual1}) demonstrates a critical failure mode. The prompt specifies a gaze-following experiment where a woman's gaze should guide a man's attention to a drawer. The resulting video, however, shows a plausible but incorrect interaction—a direct conversation with mutual eye contact. Our agent correctly diagnoses this mismatch: it recognizes the video's visual quality (D5: Video Plausibility = 1) but correctly assigns zeros to the other four dimensions because the core experiment was not performed, the prompt was not followed, and the specific social cues for gaze-following were absent.

The second case (Fig.~\ref{fig:visual2}) provides a textbook example of a successful generation. Tasked with showing instrumental helping, the video portrays a boy noticing a woman's repeated failed attempts to reach a pen and subsequently intervening to help her. This sequence perfectly aligns with the prompt and the intended social logic. As a result, the VLM judge awards a full score across all five dimensions, confirming that the model successfully rendered the prompt's narrative, the core psychological concept, coherent agent reasoning, and effective social cues within a plausible scene. These cases validate our agent's capacity to not only reward faithful and logically sound generations but also to penalize subtle yet critical failures where the underlying experimental paradigm is violated, even if the resulting video appears socially coherent on the surface.

\section{Conclusion}

We present the first benchmark dedicated to evaluating \emph{social reasoning} in video generation.
Unlike existing evaluations that focus primarily on perceptual or physical dimensions, our benchmark targets the causal, intentional, and socially grounded behaviors that underlie human interaction.
By grounding the benchmark in eight well-established components of social cognition and operationalizing them through thirty classic psychological experiments, we provide a theoretically principled and interpretable foundation for assessing social reasoning in generative models.To enable scalable and training-free benchmark construction, we introduce a four-agent pipeline that transforms abstract psychological paradigms into validated, difficulty-controlled, video-ready prompts and performs automatic evaluation through a vision–language model judge.
Our analyses demonstrate that both generation-side agents and evaluation-side agents exhibit strong alignment with human reasoning, enabling reliable large-scale assessment without manual annotation. Experiments across eight state-of-the-art video generation models reveal a substantial gap between perceptual realism and social coherence: while leading proprietary systems show emerging competence in goal understanding, joint attention, and prosocial behavior, even the strongest models fail systematically on belief-based inference, subtle cue integration, and multi-agent coordination.

{
    \small
    \bibliographystyle{ieeenat_fullname}
    \balance
    \bibliography{main}
}


\end{document}